\documentclass[letterpaper, 10 pt, conference]{ieeeconf}  

\IEEEoverridecommandlockouts                              

\usepackage{amsmath,amssymb,amsfonts}
\usepackage{algorithmic,algorithm}
\usepackage{xcolor}
\usepackage{url}
\usepackage{siunitx}
\usepackage{graphicx}
\usepackage{caption}
\usepackage{subcaption}
\usepackage{hyperref}
\usepackage{dblfloatfix}
\usepackage{multirow}
\usepackage{cite}
\usepackage{tabularray}

\graphicspath{ {./Figures/} }

\begin{document}




\title{\LARGE \bf
Leveraging Computation of Expectation Models for Commonsense Affordance Estimation on 3D Scene Graphs}

\author{Mario A.V. Saucedo, Nikolaos Stathoulopoulos, Akash Patel, Christoforos Kanellakis\\
and George Nikolakopoulos
\thanks{The authors are with Robotics \& AI Team, Department of Computer, Electrical and Space Engineering, Lule\r{a} University of Technology, Lule\r{a} SE-97187, Sweden. 
        {Corresponding author: \tt\small marval@ltu.se}}
\thanks{This work has been partially funded by the European Union's Horizon Europe Research and Innovation Programme, under the Grant Agreement No. 101119774 SPEAR.}
}

\maketitle
\thispagestyle{empty}
\pagestyle{empty}

\begin{abstract}
This article studies the commonsense object affordance concept for enabling close-to-human task planning and task optimization of embodied robotic agents in urban environments. The focus of the object affordance is on reasoning how to effectively identify object’s inherent utility during the task execution, which in this work is enabled through the analysis of contextual relations of sparse information of 3D scene graphs. The proposed framework develops a Correlation Information (CECI) model to learn probability distributions using a Graph Convolutional Network, allowing to extract the commonsense affordance for individual members of a semantic class. The overall framework was experimentally validated in a real-world indoor environment, showcasing the ability of the method to level with human commonsense.
For a video of the article, showcasing the experimental demonstration, please refer to the following link: \url{https://youtu.be/BDCMVx2GiQE}

\end{abstract}

\section{Introduction}

Achieving a level of understanding and reasoning in robots comparable to humans has proven to be a significant challenge in robotics, where one of the most common bottlenecks relates to knowledge representation. Although modern sensors like stereo cameras and 3D LiDARs enable robots to perceive the world in a manner akin to human senses, storing and processing this information onboard mobile devices proves to be significantly inefficient by comparison. 
Lately, the introduction of 3D Scene Graphs (3DSG) presented a promising solution. In this approach, the knowledge of the environment is represented in a sparse abstract graph, where nodes signify objects and edges denote the relationships among these nodes~\cite{Armeni2019,Kim2020,Wald2020,Rosinol2021,Wu2021,Hughes2022,Wald2022,room,Feng2023,Wang2023,Wu2023}. 
Thus, the adoption of 3D scene graphs has enabled a broader range of algorithms capable of approximating human-like reasoning across various aspects of robotics missions.
Some of the main examples include tasks such as navigation~\cite{ginting2023}, variability estimation~\cite{Looper2023}, object localization~\cite{Giuliari2022}, and partial scene completion~\cite{saucedo2024belief}.
Similar to humans, robots are expected to make commonsense decisions when confronted with incomplete information, thus optimizing complex tasks that typically necessitate human intervention.

To illustrate this concept, let's consider the following scenario: a robot tasked with fetching a chair in an environment where it has only partial information about its surroundings. Similar to humans, the robot is constrained by physical limitations, meaning not every chair can be feasibly moved. Therefore, the robot must assess the \textit{affordance} of each chair before proceeding with the task. While one approach would involve inspecting each chair and evaluating its \textit{affordance} individually, this would be highly inefficient.
In contrast to human behavior, where the \textit{affordance} of an object is often assumed beforehand, allowing for more streamlined planning, the robot should ideally exhibit a similar proactive approach. It should prioritize visiting locations where chairs with the highest likelihood of being easily movable are expected to be found. For instance, an office chair, known for its mobility, would be a prime target. This approach optimizes the robot's decision-making process and aligns more closely with human-like behavior.

\begin{figure*}[!ht]
    \centering
    \includegraphics[width=\textwidth]{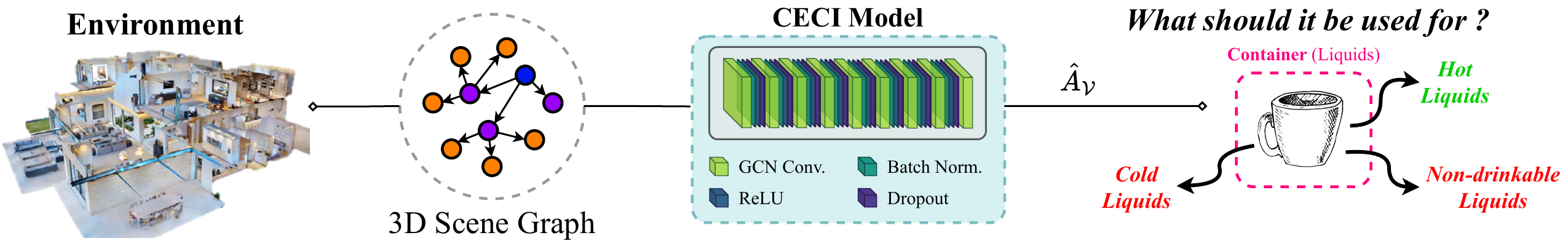}
    \caption{Depiction of the proposed affordance estimation method based on 3D scene graphs, where the environment is first abstracted into a 3D scene graph representing \textcolor[HTML]{0018EC}{building}, \textcolor[HTML]{9702FA}{rooms} and \textcolor[HTML]{FF8000}{objects}, and then input to the CECI model for affordance estimation in order to determine the set of \textcolor[HTML]{00CC00}{commonsense affordances} $\hat{A}_{\mathcal{V}}$ for individual members of a \textcolor[HTML]{FF0080}{semantic class}.}
    \label{fig:diag}
\end{figure*}

While recent works have tackled the task of semantic scene completion in 3DSG~\cite{saucedo2024belief}, the estimation of per-node affordance for members of general categories remains largely unexplored in the current state-of-the-art. To address this challenge, we leverage a novel model for Computation of Expectation based on Correlation Information (CECI) for affordance estimation on 3DSG. An overview of our proposed approach is depicted in Fig.~\ref{fig:diag}.

\subsection*{--- Contributions}




The contributions of this article are summarized as follows. The focus of this work is to extract high-level information out of a given 3D scene graph $\mathcal{G}$, in where objects are encoded as entry-level categories represented by a set $C$, by proposing an approach to meaningfully estimate the affordance of individual members of a semantic class. Towards this end, we establish the novel concept of \textit{commonsense affordance} as a utility-driven attribute of a given object in a 3D scene graph $\mathcal{G}$ which enables high-level reasoning over the properties of specific members of a semantic class.

To achieve this, we design and implement a novel Graph Convolutional Network (GCN) model for the Computation of Expectation of the commonsense affordance of any given object node $\hat{a}_i$ within a semantic class label $c_i$, based on Correlation Information (CECI) between $\hat{a}_i$ and the objects that are part of the given 3D scene graph $\mathcal{G}$. 
In order to do so, we present a novel methodology for utilizing semantically annotated real-life 3D spaces to generate and annotate a 3D scene graphs dataset with ground truth commonsense affordance. The generated dataset is then used to train the proposed CECI model and to validate the proposed methodology. Followed by an in-depth analysis of the results for the learned correlation information among the semantic class labels in $C$ and the predicted commonsense affordance~$\hat{A}$.

Finally, we present real-world experiments for commonsense affordance estimation by constructing a 3D scene graph using information gathered by a legged robotic platform through the exploration of an unknown indoor environment. The experimental results demonstrate the performance of the proposed method for estimating the commonsense affordance of individual members of an entry-level category.


\section{Related Work}

Scene graphs are typically constructed using methods such as geometric reasoning, semantic segmentation, and deep learning-based approaches. These methods involve analyzing the spatial relationships between objects in a scene, segmenting the scene into meaningful regions, and using neural networks to infer object categories and their relationships. 
In~\cite{Kim2020} the framework begins the 3D scene graph construction by preprocessing input images and extracting relative poses using techniques such as visual odometry or SLAM. It subsequently categorizes frames into key, anchor, or garbage frames to enhance processing efficiency and then employs region proposal and object recognition for object detection and categorization. Following this, it utilizes relation extraction to identify intricate object relationships before employing spurious detection rejection to filter out erroneous detections. Finally, it constructs local 3D scene graphs for each frame, which are integrated to continually update the global 3D scene graph, ensuring a comprehensive representation of the scene over time.
More recently, Hughes et al.~\cite{Hughes2022} proposed \textit{Hydra}, a framework for real-time 3D scene graph construction and optimization. The paper outlines a method for constructing 3D scene graphs in real-time, starting with the incremental creation of layers as the robot explores its environment. This involves generating a local Euclidean Signed Distance Function (ESDF), extracting a topological map of places, and segmenting these places into rooms. Additionally, the approach incorporates loop closure detection and optimization, utilizing hierarchical descriptors and embedded deformation graphs to correct all layers of the scene graph in response to loop closures.

Apart from the construction of 3D scene graphs, several studies have addressed semantic scene understanding using 3D scene graphs. For instance, Giuliari et al.~\cite{Giuliari2022} proposed the concept of a spatial commonsense graph, specifically designed to tackle the challenging task of object localization within partial 3D scans. Building upon this foundation, Looper et al. \cite{Looper2023} introduced a dynamic approach with their variable scene graph, designed to estimate semantic scene variability by discerning changes in position, semantic state, and overall scene composition. In a similar vein, Ginting et al.~\cite{ginting2023} devised a semantic belief graph aimed at navigating through environments fraught with perceptual uncertainty, offering a robust framework for semantic-based planning in extreme conditions. Finally, we can find~\cite{saucedo2024belief}, in where a novel method for semantic scene completion is presented, the authors additionally proposed a new graph representation, namely Belief Scene Graphs (BSGs), as an expansion of a 3D scene graph that incorporates a novel kind of nodes of probabilistic nature to depict the predicted expected objects (i.e. unseen objects).

Furthermore, the concept of affordance within the context of 3D scene graphs has garnered attention, as evidenced by the work of Rana et al. \cite{rana2023sayplan}. Their research introduced a comprehensive task planning framework leveraging 3D scene graphs, incorporating an iterative re-planning pipeline. This approach not only rectifies infeasible actions but also mitigates planning failures, underscoring the significance of affordance considerations within the realm of semantic scene understanding.


In contrast to previous works, we present a method to estimate the novel concept of commonsense affordance for individual members of a semantic class. The proposed methods consist of a novel CECI model architecture~\cite{saucedo2024belief} that allows to determine the commonsense affordance of the different objects present in a 3D scene graph. 

\section{Problem Formulation}
\label{sec:pd}

\textbf{Commonsense Affordance:} In any given 3D scene graph $\mathcal{G}^\prime$ the set of nodes $\mathcal{V}$ will present a single attribute consisting of its respective labeled object class. 
This allows to estimate object affordance on a category-level (e.g. container for liquids), under the assumption that objects of the same class share, although to a different degree, a similar set of affordances $A$. Nevertheless, in order to imitate human behavior, affordance is often required in a sub-category-level (e.g. cup). This sub-category affordance is referred to as commonsense affordance $A^*$ for the rest of the article. Commonsense affordance is often grounded in a deeper understanding of the purpose that specific objects of a broad category are designed for, and as such implies a stronger affordance for task-specific actions.

\textbf{Semantic Class Labels:} The selection of object categories in object detection datasets is a non-trivial exercise. The categories must form a representative set of all categories, be relevant to practical applications, and occur with high enough frequency to enable the collection of a large dataset.
To enable the practical collection of a significant number of instances per category, most datasets are limited to entry-level category labels that are commonly used by humans when describing objects \cite{cocodataset}.
Similarly, most 3DSG frameworks model each object as a node and node attributes containing its semantic class label $c_i$. Furthermore, rooms are often also labeled as a single semantic class in order to allow generalization to different environments~\cite{Rosinol2021,Hughes2022,room,saucedo2024belief}.
This means that in most cases we don't count with the sub-class label $c_i^*$ of any given object, therefore we can not determine in a direct manner its corresponding commonsense affordance $a_i^*$. 
Instead, in this article, we look to find a reasonably approximate set of probability distributions $\hat{A}$ for the set of commonsense affordances $A^*$ based on the semantic class labels of the nodes in the graph $\mathcal{G}^\prime$.

\textbf{Modelling Affordance:}
We need to estimate the commonsense affordance of a given node $\gamma \in \mathcal{V}$, defined as the set of affordance probabilities $\hat{A}_{\gamma} = \{\hat{a}_1, ..., \hat{a}_n\}$ with $n \in \mathbb{Z^+}$.
The probability of any given affordance $\hat{a}_i$ can be modeled as the conditional probability among the affordance $\hat{a}_i$ and a set of $m \in \mathbb{Z^+}$ objects and room observations $B = \{b_1, ..., b_m\}$, which are assumed to be independent of each other. Then the expectation for the affordance $\hat{a}_i$ is given by: 
\begin{equation} 
\boldsymbol{P(\hat{a}_i|B)} = \sum_{j = 1}^m P(\hat{a}_i \cap b_j) \Big/ \prod_{j=1}^m P(b_j)
\label{eqn:1} 
\end{equation}
Note that the affordance $\hat{a}_i$ and observation $b_j$ are being treated as singleton events with associated probabilities. 
Likewise, we can estimate the set of probabilities $\hat{A}_{\gamma}$ for every commonsense affordance:
\begin{equation} 
\boldsymbol{\hat{A}_{\gamma} =} 
\begin{pmatrix}
P(\hat{a}_1|B & P(\hat{a}_2|B) & \cdots & P(\hat{a}_n|B)
\end{pmatrix}
\label{eqn:2} 
\end{equation}

\textbf{Computation of Expectation based on Correlation Information:} 
Let $\hat{A}_{\mathcal{V}}$ denote the real set of conditional probabilities with respect to all the nodes $\mathcal{V}$ of a given 3D scene graph $\mathcal{G}^\prime$. In order to estimate $\hat{A}_{\mathcal{V}}$ and to allow the scalability of the framework, we propose a novel Graph Convolutional Networks (GCN) inspired by the Computation of Expectation based on Correlation Information (CECI) model proposed in \cite{saucedo2024belief}. The novel CECI model allows to estimate the commonsense affordances of any given 3D scene graph based on the sparse information of the graph's topology.

\section{Commonsense Affordance}

This Section details the design of the novel CECI model for commonsense affordance estimation, as well as the methodology used to develop the dataset.

\subsection{CECI Model Selection and Design}
\label{sec:ceci}

CECI models are structured around the idea of estimating probability distributions rather than precise semantic labels \cite{saucedo2024belief}. This particular property allows us to classify nodes into soft-abstract categories rather than hard-defined ones, which closer resembles human knowledge. The abstract nature of the output also allows to distinguish among the different objects of the same semantic class based on their commonsense affordance. In more detail, the proposed CECI model consists of 9 GCN convolutional layers~\cite{gcn}, followed by batch normalization~\cite{batchn}, 
ReLU and dropout. The input to the network is a given 3DSG, denoted as $\mathcal{G}^\prime = (\mathcal{V},\mathcal{E})$ and composed of the set of directional edges $\mathcal{E}$ and the set of vertex $\mathcal{V}$.
The node attribute for any given node $\gamma \in \mathcal{V}$ consists of the semantic class label $c_i \in C$, where $C = \{c_1, ..., c_k\}$ with $k \in \mathbb{Z^+}$ denote the set of semantic class labels present in the graph $\mathcal{G}^\prime$. The output is the predicted set of probabilities for commonsense affordances $\hat{A}_{\mathcal{V}}$, while the  
overall network architecture is visualized in Fig.~\ref{fig:diag}.

\subsection{Dataset Generation and Data Augmentation}

This article presents for the first time in the current state-of-the-art the task of commonsense affordance estimation, meaning that currently there are no available datasets for the training of the proposed CECI model.
Instead, we used the Habitat-Matterport 3D Research Dataset (HM3D) \cite{hm3d}, currently the world's largest dataset of real-life residential, commercial, and civic spaces, as the base to generate the data needed for training. 
Initially, we generated a custom-made mapping that involves the grouping and filtering of the 1659 semantic categories present on the dataset, into a subset of 45 hand-picked semantic class labels. The criteria for the selection of the labels consisted of three metrics: (i) their frequency on the scenes, (ii) their relevance to a robotic task, and (iii) the difference between their category-level affordance $A$ and the commonsense affordance $A^*$ of the available sub-categories. Furthermore, we have chosen to delimit the original set of room labels to a single general class label (i.e. ``room") to match the information present in a Belief Scene Graph~\cite{saucedo2024belief} and most 3DSG~\cite{Rosinol2021,Hughes2022,room}. 

Secondly, for each of the dataset's semantically annotated 3D spaces, we constructed a ground truth 3DSG $\mathcal{G} = (\mathcal{V},\mathcal{E})$ with $\mathcal{L} = \{Building,Rooms,Objects\}$, a set of semantic layers utilized in previous works \cite{Armeni2019,Rosinol2021,Hughes2022}. The nodes consist of two attributes, the first one being the semantic class label and the second one being the set of ground truth commonsense affordances $A^*$, while the edges represent descendant relationships (i.e. the child node is physically contained on the parent node).
Then, for each ground truth graph $\mathcal{G}$ we apply the augmentation process presented in \cite{saucedo2024belief}, where we generated in addition, a series of incomplete graphs $\mathcal{G}^\prime$ by deleting object nodes at random, until the number of deleted nodes is equal to 20\% of the original number of nodes. This steps helps to increase the diversity of the dataset and avoid over-fitting.
Finally, we delete the ground truth affordance from the node attributes of the input graphs $\mathcal{G}^\prime$. 

\subsection{Ground Truth Commonsense Affordance}

Translating the intuitive notion of commonsense affordance into a numerical representation presents multiple challenges. Although in Section \ref{sec:pd} we proposed a vectorized representation of this concept, in real-life it is extremely challenging to compute such probabilities due to the heavy data requirements for this purpose. While the model proposed in Section \ref{sec:ceci} will allow us to approximate these values, the need for ground truth data for the training process remains.
In order to generate the ground truth data, we use human-determined affordances for each available pre-annotated subgroup of objects on the dataset that belongs to a broader category group. In other words, in the graph we use the entry-level semantic class label (e.g. chair) for describing a broad group of objects, due to the reasons described in Section \ref{sec:pd}, but in the dataset we have access to human-annotated subgroups that belong to this broader category (e.g. office chair, sofa chair, etc.). 

The full process to generate the ground truth commonsense affordance goes as follows: For each entry-label category we generate a base affordance set $A_{chair} = \{{carried}, {dragged}, {stepped}\}$, which denotes a list of possible affordances for the semantic class labels present in the graph $\mathcal{G}$. Afterward, human annotators are tasked to select which affordances correspond to the commonsense affordance vector of each sub-category, for example $A_{chair}^{office-chair} = [0,1,0]$. This implies that despite the subgroup \textit{office chair} sharing the general affordance of $carried$ with other members of the group \textit{chair}, under human commonsense, it should not be carried but instead dragged. Finally, the resulting commonsense affordance vector is normalized and represents the ground truth probability distribution of the sub-category to be learned by the proposed CECI model.





\section{Experimental Evaluation}
In this Section, we first provide the training parameters of the CECI model, followed by the validation using Wasserstein distance 
and energy distance. 
Secondly, we present an in-depth analysis of the leaned correlation information among the set of semantic class labels $C$ and the estimated commonsense affordances $\hat{A}$. Finally, we report the results for commonsense affordance estimation in a real-world indoor environment with a Boston~Dynamics Spot~Legged~Robot.

\subsection{CECI Model Training}
In the developed implementation, we used a total of 45 labeled object classes. The generated dataset was used in an 80\% / 10\% / 10\% split for training, validation and testing respectively.
The training consisted of 5000 epochs with a batch size of 50. We used an Adam optimizer \cite{adam} with a learning rate of 0.01 and a learning rate decay of 5$e^{-6}$. 
The loss function was chosen to be a Mean Squared Error (MSE).

\subsection{Validation Metrics}
The validation process considered two main metrics for the statistical distance between the probability distributions of the estimated commonsense affordance and the ground truth. 
The computed metrics were the Wasserstein distance (i.e. the earthmover’s distance) 
and the energy distance 
for the pair of probabilistic distributions, $\hat{A}$ for the prediction and $A^*$ for the ground truth. Table~\ref{table:distances} shows the mean, variance, skewness and kurtosis for each of the computed distances. Overall, the predicted distribution is fairly similar to the ground truth, it can be observed that both metrics presented a low mean and a low variance.

\begin{table}[!htbp]
\centering
\caption{Statistical Distance}
\label{table:distances}
\resizebox{\linewidth}{!}{%
\begin{tblr}{
  width = \linewidth,
  colspec = {Q[177]Q[183]Q[198]Q[183]Q[198]},
  cells = {c},
  vline{2} = {-}{},
  hline{1-2,4} = {-}{},
  hline{2} = {2}{-}{},
}
\textbf{Metric} & \textbf{Mean} & \textbf{Variance} & \textbf{Skewness} & \textbf{Kurtosis} \\
Wasserstein     & $0.1517$      & $0.01371$         & $0.5635$          & $-0.9086$         \\
Energy          & $0.3205$      & $0.02245$         & $0.0491$          & $-0.8878$         
\end{tblr}
}
\end{table}


Furthermore, we compute the correlation among the commonsense affordances of 4 different entry-level categories against the 45 semantic class labels present in the proposed dataset.
The showcased entry-level categories are those that present the biggest gap between the commonsense affordances of their respective sub-categories.
The computed correlations are then used to estimate the Frobenius norm 
of the difference between the correlations of the predicted commonsense affordances and the ground truth data. 
The results are presented in Table \ref{table:corr} and indicate a fair degree of similarity among the correlations of the predicted commonsense affordances and the ground truth.

\begin{table}[ht!]
\centering
\caption{Correlation Comparison}
\label{table:corr}
\resizebox{\linewidth}{!}{%
\begin{tblr}{
  width = \linewidth,
  colspec = {Q[274]Q[142]Q[131]Q[167]Q[209]},
  row{odd} = {c},
  cell{1}{2} = {c=3}{0.44\linewidth},
  cell{2}{1} = {r=2}{},
  cell{2}{2} = {c=3}{0.44\linewidth,c},
  cell{2}{5} = {r=2}{c},
  cell{4}{1} = {r=2}{},
  cell{4}{2} = {c=3}{0.44\linewidth,c},
  cell{4}{5} = {r=2}{c},
  cell{6}{1} = {r=2}{},
  cell{6}{2} = {c=3}{0.44\linewidth,c},
  cell{6}{5} = {r=2}{c},
  cell{8}{1} = {r=2}{},
  cell{8}{2} = {c=3}{0.44\linewidth,c},
  cell{8}{5} = {r=2}{c},
  vline{2-3} = {1-2,3,4,5,6,7,8,9}{},
  vline{3-5} = {1,2,3,4,5,6,7,8,9}{},
  hline{1-2,4,6,8,10} = {-}{},
  hline{2} = {2}{-}{},
  hline{3,5,7,9} = {2-4}{},
}
\textbf{Semantic Class Label} & \textbf{Commonsense Affordance}           &             &               & \textbf{Frobenius Norm} \\
Chair                         & \textit{What should I do with it ?}       &             &               & 0.0605                  \\
                              & Dragged                                   & Carried     & Stepped       &                         \\
Fabric                        & \textit{What should be covered with it ?} &             &               & 0.0606                  \\
                              & Floor                                     & Furniture   & Wet-surface   &                         \\
Container\qquad Solids              & \textit{What should be stored on it ?}    &             &               & 0.2062                  \\
                              & Food                                      & Garbage     & Clothes       &                         \\
Container Liquids             & \textit{What should be poured on it ?}    &             &               & 0.1697                  \\
                              & Cold-liquids                              & Hot-liquids & Non-drinkable &                         
\end{tblr}
}
\end{table}


\begin{figure*}[!ht]
    \begin{minipage}{0.93\textwidth}%
        \begin{minipage}{0.5\textwidth}%
            \begin{subfigure}{\linewidth}%
                \includegraphics[width=0.95\linewidth]{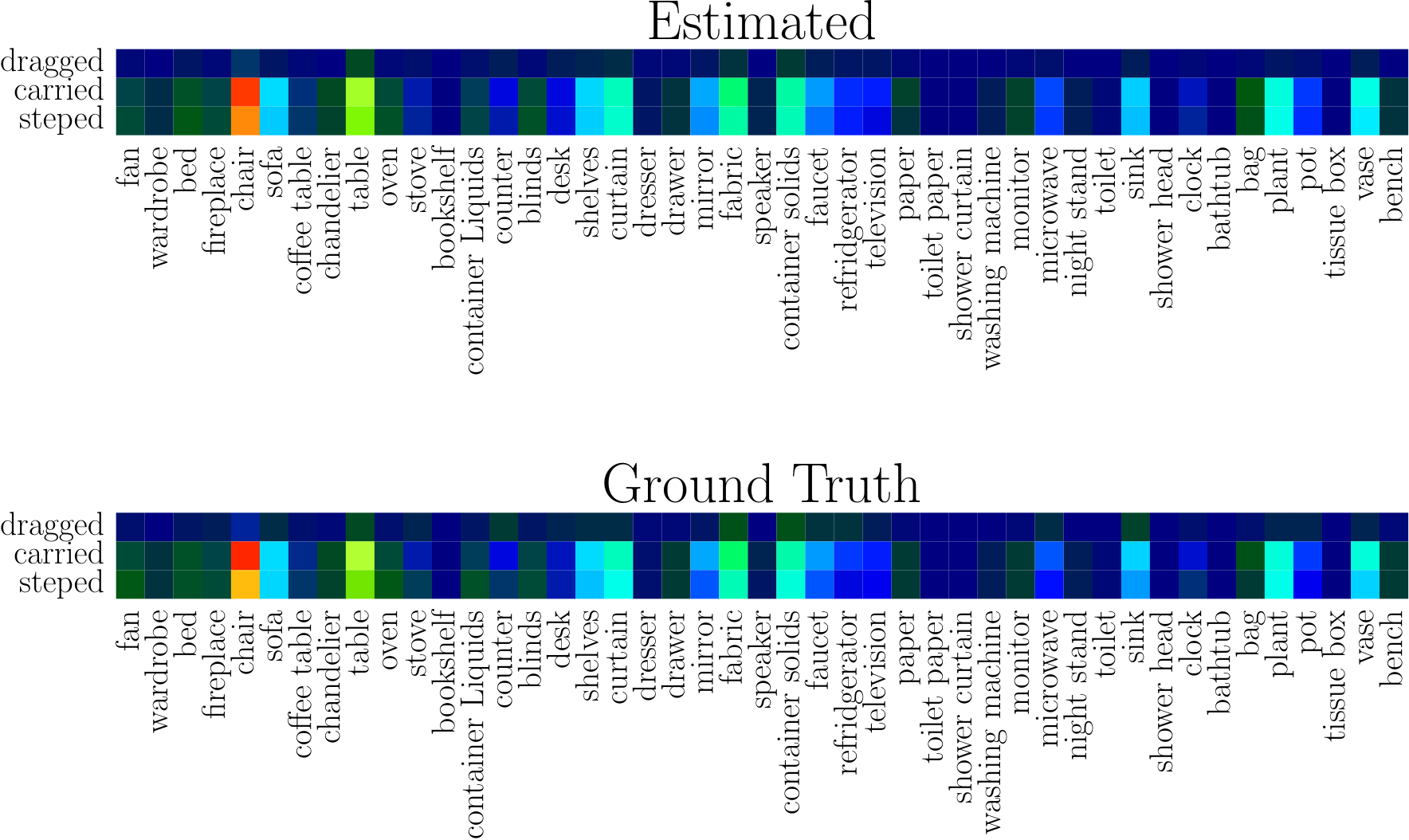}
                \caption{Chair} 
                \label{fig:case-a}
            \end{subfigure}
        \end{minipage}
    \vspace{2mm}
        \begin{minipage}{0.5\textwidth}%
            \begin{subfigure}{\linewidth}%
                \includegraphics[width=0.95\linewidth]{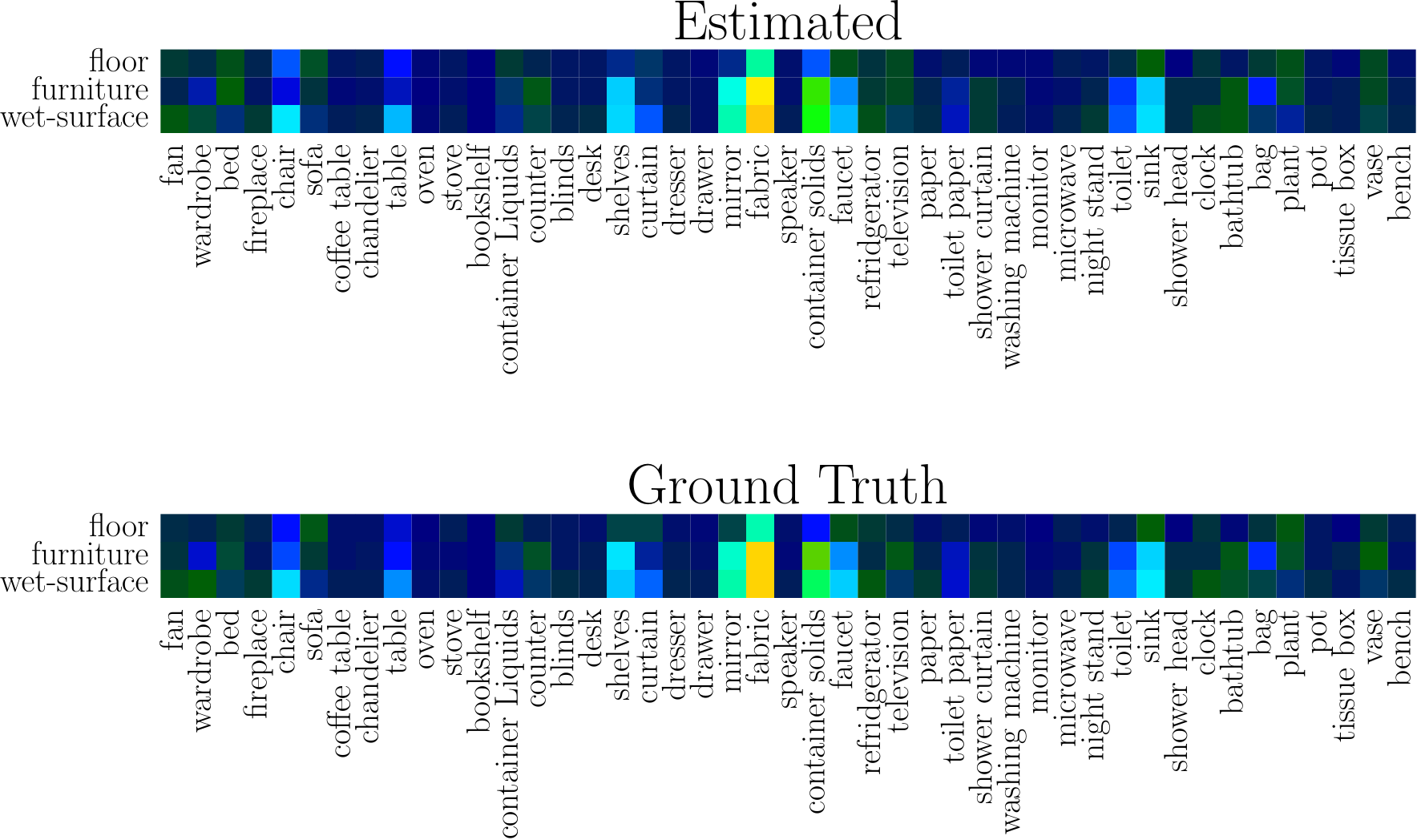}
                \caption{Fabric} 
                \label{fig:case-b}
            \end{subfigure}
        \end{minipage}
        \begin{minipage}{0.5\textwidth}%
            \begin{subfigure}{\linewidth}%
                \includegraphics[width=0.95\linewidth]{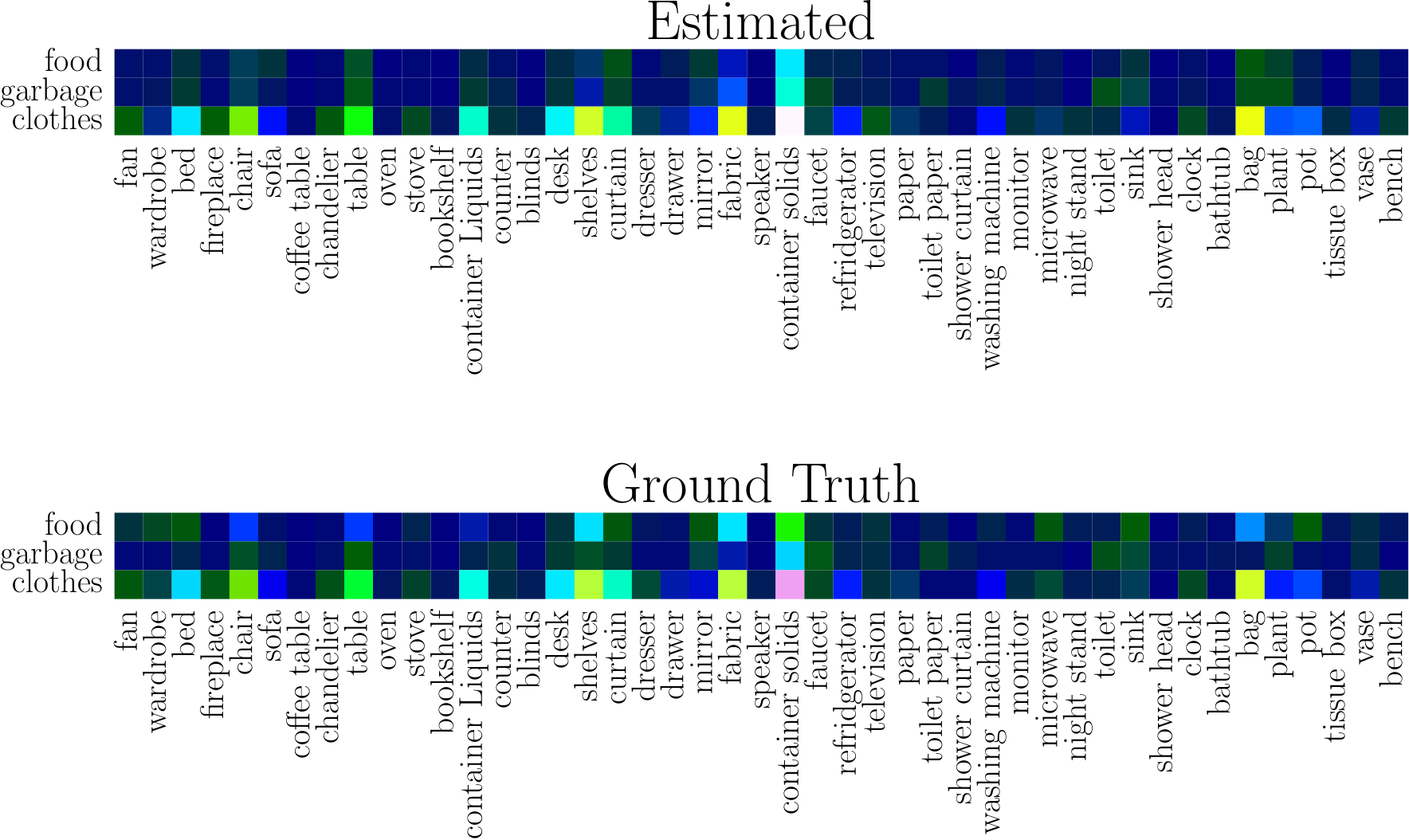}
                \caption{Container Solids} 
                \label{fig:case-c}
            \end{subfigure}
        \end{minipage}
        \begin{minipage}{0.5\textwidth}%
            \begin{subfigure}{\linewidth}%
                \includegraphics[width=0.95\linewidth]{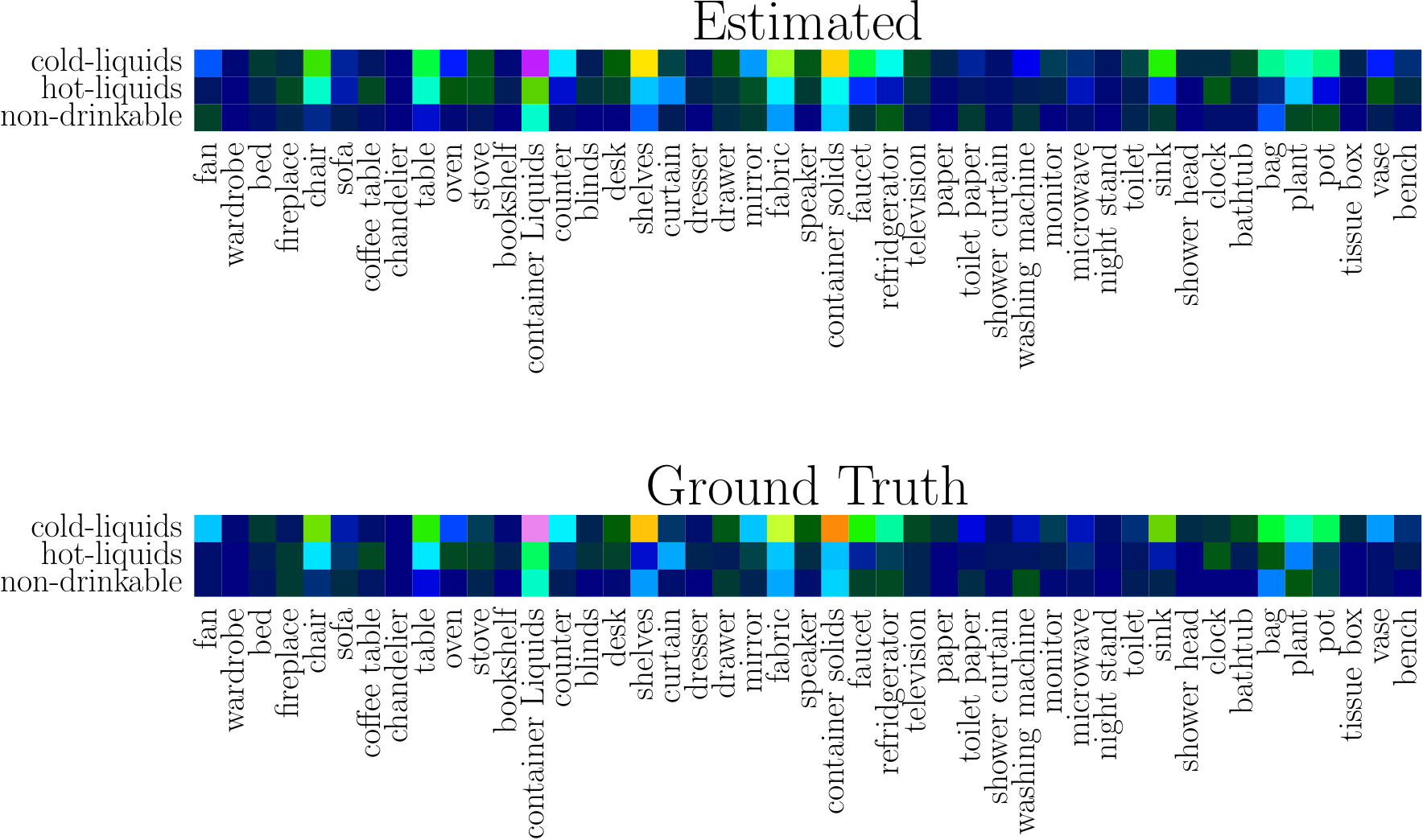}
                \caption{Container Liquids} 
                \label{fig:case-d}
            \end{subfigure}
        \end{minipage}
    \end{minipage}
    \hspace{0.5mm}
    \begin{minipage}{0.05\textwidth}%
    \begin{subfigure}{\linewidth}%
        \includegraphics[width=\textwidth]{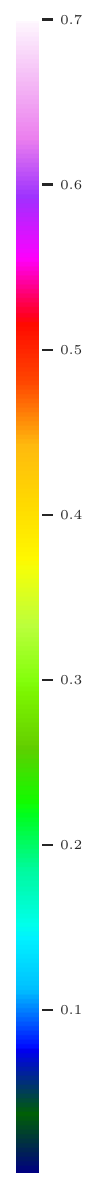}
        \label{fig:merge-scenario}
    \end{subfigure}
    \end{minipage}%
    \caption{Depiction of the computed correlations among the 45 semantic class labels present in the generated dataset and the predicted commonsense affordances for 4 entry-level categories. The correlations of the ground truth data are present at the bottom for comparison.}
    \label{fig:corr}
    \vspace{-0.5em}
\end{figure*}

\subsection{Qualitative Analysis}

Figure \ref{fig:corr} depicts the computed correlations and showcases the performance of the CECI model for learning the underlying correlations between the different sub-categories semantic class labels present in the generated dataset.
As can be seen, the strongest correlation of any affordance is with the respective entry-level category it belongs to. So that if we add the correlation values of the three corresponding affordances of a semantic class, we will obtain a correlation of 1 (i.e. maximum correlation). 
Moreover, we can observe a high correlation among objects, like \textit{table} and \textit{sofa} with the affordance $carried$ of the \textit{chair} class. Likewise, for the affordance $wet-surfaces$ of the class \textit{fabric} with respect to the objects \textit{mirror} and \textit{sink}. 
On the other hand, for the class \textit{container solids} we have a strong correlation for the objects \textit{bed} and \textit{washing machine} with respect to the affordance $clothes$ of the class \textit{container solids}. Similarly, the affordance $cold-liquids$ of the class \textit{container liquids} presents a high correlation with the objects \textit{table} and \textit{refrigerator}.



\subsection{Field Test Results for Commonsense Affordance}

\begin{figure}[!ht]
        \centering
        \includegraphics[width=\columnwidth]{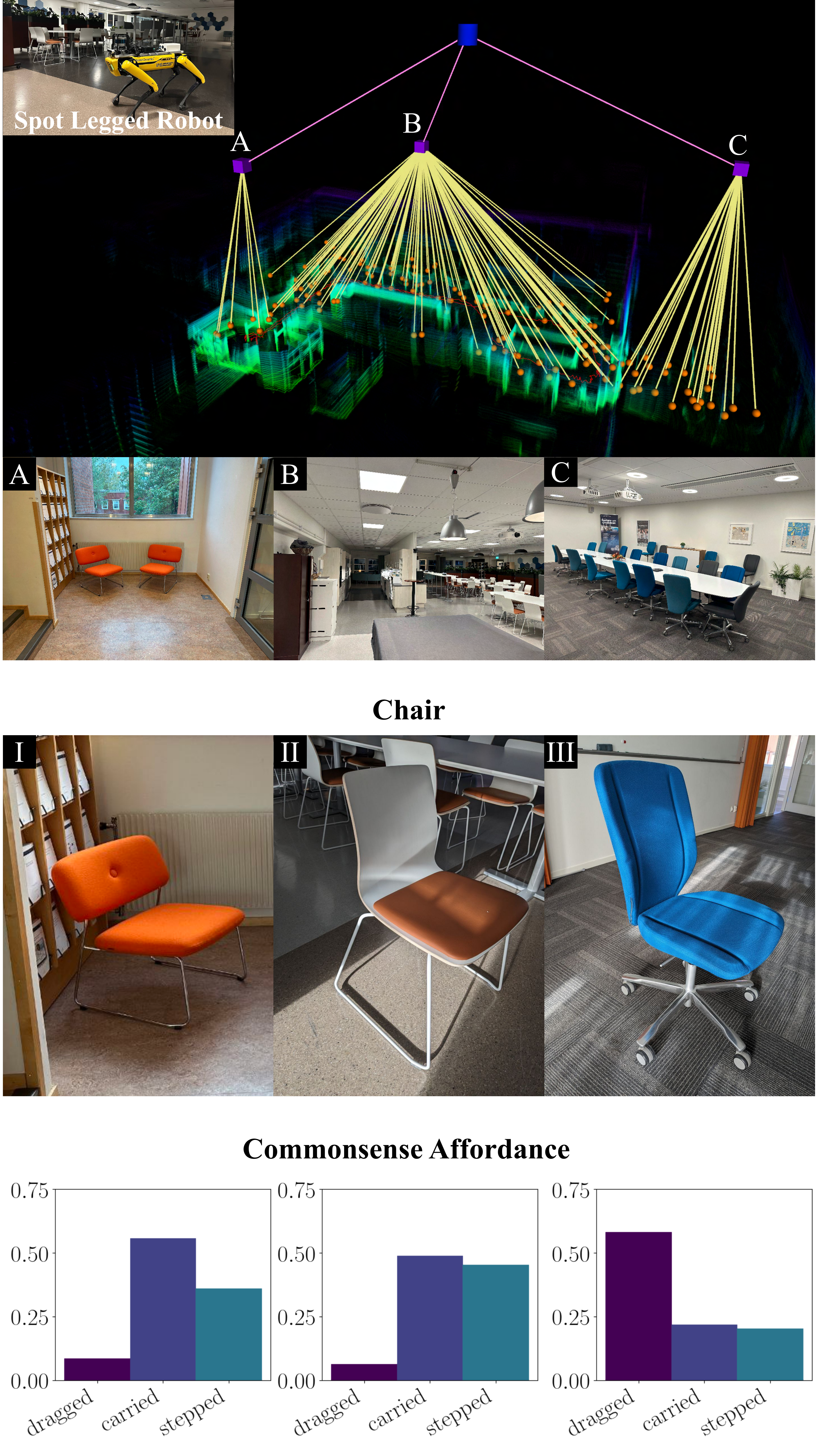}
        \caption{Depiction of the proposed affordance estimation method, where the environment is represented by a 3D scene graph with \textcolor[HTML]{0018EC}{building}, \textcolor[HTML]{9702FA}{rooms} and \textcolor[HTML]{FF8000}{objects}, and used to determine the set of commonsense affordances $\hat{A}$ for the individual member of the same semantic class (i.e. \textit{chair}).}
        \label{fig:fika}
        \vspace{-1em}
    \end{figure}

We tested our approach on the Boston Dynamics Spot legged robot in a real indoors environment 
with the goal to contrast the estimated commonsense affordance with human commonsense. The environment consisted of a small waiting room (i.e. Room A), a shared kitchen with a dining area (i.e. Room B), and a meeting room (i.e. Room C). 
The experiment required for the robot to traverse the environment while generating a 3D scene graph. The graph was then enhanced using the proposed CECI model to estimate the commonsense affordance of the objects present in the environment. 
Figure \ref{fig:fika} presents the generated 3D scene graph for one of the experimental trials while depicting alongside images of each of the traversed rooms, the semantic class of interest (i.e. chair), and the estimated commonsense affordance.

This experiment helps to illustrate the advantages of the proposed method for semantic scene understanding.
Rather than relying on an object detection model trained on a dataset of densely annotated sub-categories, we can use a more compact and fast model trained on entry-level categories.
This feature proves to be increasingly helpful when working with seldom environments (e.g. subterranean) where the collection of large datasets is a challenging task.
Furthermore, the probability distributions for the set of commonsense affordances presented at the bottom of Fig. \ref{fig:fika} help to illustrate the performance of the proposed method.
Furthermore, it can be observed how different instances of the same semantic class present different distributions for their commonsense affordance based on the nature of the surrounding objects. 


\section{Conclusions}

A novel concept of \textit{commonsense affordance} was introduced in this work, as a utility-driven attribute of a given object in a 3D scene graph. Reasoning about the commonsense affordance of objects through the sparse information in a 3D scene graph represents a crucial incremental step in the way robotic systems understand their environment, enabling close-to-human task planning and task optimization. The 
proposed CECI model allows to learn probability distributions throughout a GCN and enables the translation of an intuitive representation of commonsense affordance into a computational method. 
The overall framework was experimentally validated in a real-world indoor environment, showcasing the ability of the method to level with human commonsense.
Future works will focus on the implementation of the proposed framework on a broad range of robotic applications like task planning and task allocation for multi-agent scenarios.










\bibliographystyle{ieeetr}
\bibliography{References}

\end{document}